\documentclass[11pt]{article}
\usepackage{eacl2017}
\usepackage{times}
\usepackage{url}
\usepackage{latexsym}
\usepackage{amsmath}
\usepackage{amssymb}
\usepackage{graphicx}
\usepackage{caption}
\usepackage{subcaption}
\usepackage{alltt}
\usepackage{ctable}
\usepackage{multirow}
\usepackage{dsfont}
\usepackage{placeins}

\eaclfinalcopy 
\def\eaclpaperid{110} 

\def\todo#1{\bgroup {(#1)}\egroup}

\title{Unsupervised Dialogue Act Induction using Gaussian Mixtures} 

\author{Tom\'{a}\v{s} Brychc\'{i}n \\
NTIS -- New Technologies\\
 for the Information Society, \\
Faculty of Applied Sciences, \\
University of West Bohemia, \\
Czech Republic \\
  {\tt brychcin@kiv.zcu.cz} \\\And
  Pavel Kr\'{a}l \\
Department of Computer \\
Science and Engineering, \\
Faculty of Applied Sciences, \\
University of West Bohemia, \\
Czech Republic \\
  {\tt pkral@kiv.zcu.cz} \\}

\date{}

\begin{document}
\maketitle

\begin{abstract}
This paper introduces a new unsupervised approach for dialogue act induction. Given the sequence of dialogue utterances, the task is to assign them the labels representing their function in the dialogue.

Utterances are represented as real-valued vectors encoding their meaning. We model the dialogue as Hidden Markov model with emission probabilities estimated by Gaussian mixtures. We use Gibbs sampling for posterior inference.

We present the results on the standard Switchboard-DAMSL corpus. Our algorithm achieves promising results compared with strong supervised baselines and outperforms other unsupervised algorithms.

\end{abstract}

\section{Introduction}

Modeling the discourse structure is the important step toward understanding a dialogue. The description of the discourse structure is still an open issue. However, some low level characteristics have already been clearly identified, e.g. to determine the dialogue acts (DAs) \cite{Jurafsky:2009}. DA represents the meaning of an utterance in the context of the full dialogue. 

Automatic DA recognition is fundamental for many applications, starting with dialogue systems \cite{Allen:2007}. The expansion of social media in the last years has led to many other interesting applications, e.g. thread discourse structure prediction \cite{wang-EtAl:2011}, forum search \cite{Seo:2009}, or interpersonal relationship identification \cite{Diehl:2007}.

Supervised approaches to DA recognition have been successfully investigated by many authors \cite{Stolcke00,kluwer2010using,kalchbrenner-blunsom:2013}. However, annotating training data is both slow and expensive process. The expenses are increased if we consider different languages and different methods of communication (e.g. telephone conversations, e-mails, chats, forums, Facebook, Twitter, etc.). As the social media and other communication channels grow it has become crucial to investigate unsupervised models. 
There are, however, only very few related works. 

\newcite{crook2009unsupervised} use Chinese restaurant process and Gibbs sampling to cluster the utterances into flexible number of groups representing DAs in a travel-planning domain. The model lacks structural information (dependencies between DAs) and works only on the surface level (it represents an utterance as a word frequency histogram).

Sequential behavior of DAs is examined in \cite{ritter2010unsupervised}, where block Hidden Markov model (HMM) is applied to model conversations on Twitter. Authors incorporate a topic model on the top of HMM to distinguish DAs from topical clusters. They do not directly compare the resulting DAs to gold data. Instead, they measure the prediction ability of the model to estimate the order of tweets in conversation. \newcite{joty2011unsupervised} extend this work by enriching the emission distribution in HMM to also include the information about speaker and its relative position. 
A similar approach is investigated by \newcite{paul2012mixed}. They use mixed-membership Markov model which includes the functionality of topic models and assigns a latent class to each individual token in the utterance. They evaluate on the thread reconstruction task and on DA induction task,  outperforming the method of \newcite{ritter2010unsupervised}.

In this paper, we introduce a new approach to unsupervised DA induction. Similarly to previous works, it is based on HMMs to model the structural dependencies between utterances. The main novelty is the use of Multivariate Gaussian distribution for emissions (utterances) in HMM. Our approach allows to represent the utterances as real-valued vectors. It opens up opportunities to design various features encoding properties of each utterance without any modification of the proposed model. 
We evaluate our model together with several baselines (both with and without supervision) on the standard Switchboard-DAMSL corpus \cite{Jurafsky97a} and directly compare them with the human annotations.

The rest of the paper is organized as follows.
We start with the definition of our model (Sections \ref{sec:methods}, \ref{sec:inference}, and \ref{sec:features}).
We present experimental results in Section \ref{sec:xps}.
We conclude in Section \ref{sec:conclusion} and offer some directions for future work.

\section{Proposed Model\label{sec:methods}}

\begin{figure}
\centering
\includegraphics[width=0.49\textwidth]{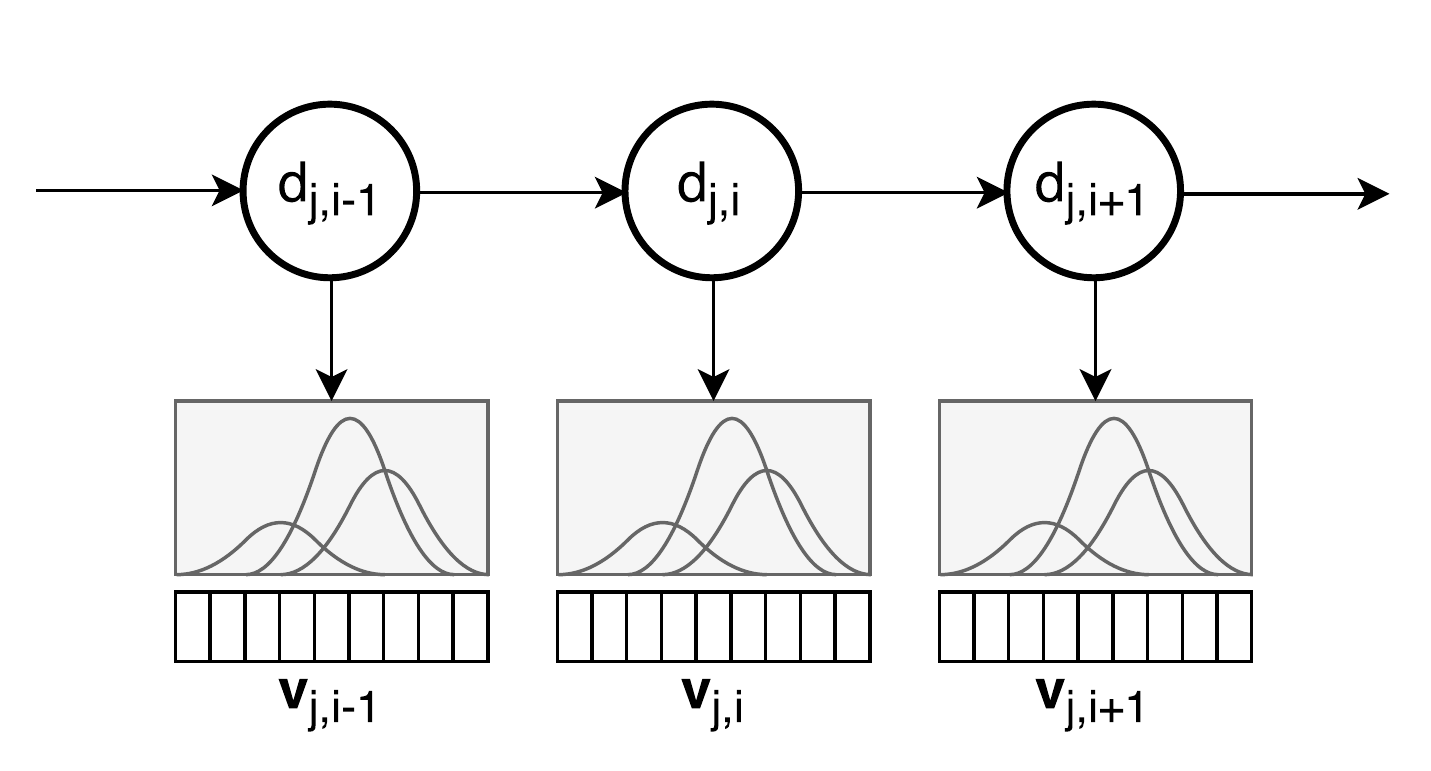}
\caption{\label{fig:model} DA model based on Gaussian mixtures.}
\end{figure}

Assume we have a set of dialogues $\boldsymbol D$. Each dialogue $\boldsymbol d_j \in \boldsymbol D$ is a sequence of DA utterances $\boldsymbol d_j = \{d_{j,i}\}_{i=1}^{N_j}$, where $N_j$ denote the length of the sequence $\boldsymbol d_j$. Let $N$ denote the length of corpora $N = \sum_{\boldsymbol d_j \in \boldsymbol D}{N_j}$. We model dialogue by HMM with $K$ discrete states representing DAs (see Figure \ref{fig:model}). 
The observation on the states is a feature vector $\boldsymbol v_{j,i}\in \mathbb{R}^M$ representing DA utterance $d_{j,i}$ (feature representation is described in Section \ref{sec:features}). HMMs thus define the following joint distribution over observations $\boldsymbol v_{j,i}$ and states $d_{j,i}$:

\begin{equation}\label{eq:posterior}
p(\boldsymbol D, \boldsymbol V) = \prod_{\boldsymbol d_j \in \boldsymbol D} {\prod_{i=1}^{N_j} {p(\boldsymbol v_{j,i}|d_{j,i})p(d_{j,i}|d_{j,i-1})}}.
\end{equation}

\noindent
Analogously to $\boldsymbol D$, $\boldsymbol V$ is a set of vector sequences $\boldsymbol v_j = \{\boldsymbol v_{j,i}\}_{i=1}^{N_j}$.

We can represent dependency between consecutive HMM states with a set of $K$ multinomial distributions $\boldsymbol \theta$ over $K$ states, such that $P(d_{j,i}|d_{j,i-1}) = \theta_{d_{j,i-1},d_{j,i}}$. We assume the probabilities $p(\boldsymbol v_{j,i}|d_{j,i})$ have the form of Multivariate Gaussian distribution with the mean $\boldsymbol\mu_{d_{j,i}}$ and covariance matrix $\boldsymbol\Sigma_{d_{j,i}}$. We place conjugate priors on parameters $\boldsymbol\mu_{d_{j,i}}$, $\boldsymbol\Sigma_{d_{j,i}}$, and $\boldsymbol \theta_{d_{j,i-1}}$: multivariate Gaussian centered at zero for the mean, an inverse-Wishart distribution for the covariance matrix, and symmetric Dirichlet prior for multinomials. We do not place any assumption on the length of the dialogue $N_j$. The full generative process can thus be summarized as follows:

\begin{enumerate}
\item For each DA $1 \le k \le K$ draw:

\begin{enumerate}
\item covariance matrix $\boldsymbol\Sigma_k \sim \mathcal{W}^{-1}(\boldsymbol \Psi, \nu)$,
\item mean vector $\boldsymbol\mu_k \sim \mathcal{N}(\boldsymbol\mu, \frac{1}{\kappa}\boldsymbol\Sigma_k)$,
\item distribution over following DAs $\boldsymbol\theta_k \sim Dir(\alpha)$.
\end{enumerate}

\item For each dialogue $\boldsymbol d_j \in \boldsymbol D$ and for each position $1 \le i \le N_j$ draw:
\begin{enumerate}
\item DA $d_{j,i} \sim Discrete(\boldsymbol \theta_{d_{j,i-1}})$,
\item feature vector $\boldsymbol v_{j,i} \sim \mathcal{N}(\boldsymbol\mu_{d_{j,i}}, \boldsymbol\Sigma_{d_{j,i}})$.
\end{enumerate}
\end{enumerate}

Note that $\kappa$ and $\nu$ represents the strength of the prior for the mean and the covariance, respectively. $\boldsymbol \Psi$ is the scale matrix of inverse-Wishart distribution.

\section{Posterior Inference\label{sec:inference}}

Our goal is to estimate the parameters of the model in a way that maximizes the joint probability in Equation \ref{eq:posterior}. We apply Gibbs sampling and gradually resample DA assignments to individual DA utterances. For doing so, we need to determine the posterior predictive distribution.

The predictive distribution of Dirichlet-multinomial has the form of additive smoothing that is well known in the context of language modeling. The hyper-parameter of Dirichlet prior determine how much is the predictive distribution smoothed. Note that we use symmetrical  Dirichlet prior so $\alpha$  in the following equations is a scalar. The predictive distribution for transitions in HMM can be expressed as

\begin{equation}
P(d_{j,i}|d_{j,i-1},\boldsymbol d_{\setminus j,i} ) = \frac{{n_{\setminus {j,i}}^{(d_{j,i}| d_{j,i-1})} + \alpha }}{{n_{\setminus {j,i}}^{(\bullet|d_{j,i-1})} +  K \alpha  }},
\end{equation}

\noindent
where $n_{\setminus {j,i}}^{(d_{j,i}| d_{j,i-1})}$ is the number of times DA $d_{j,i}$ followed DA $d_{j,i-1}$. The notation $\setminus j,i$ means to exclude the position $i$ in the $j$-th dialogue. The symbol $\bullet$ represents any DA so that $n_{\setminus j,i}^{(\bullet|d_{j,i-1})} = \sum_{1 \le k \le K} n_{\setminus {j,i}}^{(k|d_{j,i-1})}$.

The predictive distribution of Normal-inverse-Wishart distribution has the form of multivariate student $t$-distribution $t_{\nu'}(\boldsymbol v|\boldsymbol\mu',\boldsymbol\Sigma')$ with $\nu'$ degrees of freedom, mean vector $\boldsymbol\mu'$, and covariance matrix $\boldsymbol\Sigma'$. According to \cite{Murphy:2012} the parameters for posterior predictive distribution can be estimated as

\noindent
\begin{equation*}
\begin{array}{ll}
\kappa_k = \kappa + n^{(k)}, & \quad\quad\nu_k = \nu + n^{(k)},  \\
\end{array}
\end{equation*}
\begin{equation*}
\boldsymbol\Psi_k = \boldsymbol\Psi + \boldsymbol S_k + \frac{\kappa n^{(k)}}{\kappa_k} (\bar{\boldsymbol V}^{(k)} - \boldsymbol\mu)(\bar{\boldsymbol V}^{(k)} - \boldsymbol\mu)^\top,
\end{equation*}
\begin{equation}
\begin{array}{ll}
\boldsymbol\mu_k = \frac{\kappa\boldsymbol\mu +  n^{(k)} \bar{\boldsymbol V}^{(k)}}{\kappa_k}, & \quad\quad\boldsymbol\Sigma_k = \frac{\boldsymbol\Psi_k}{\nu_k - K +1},
\end{array}
\end{equation}

\noindent
where $n^{(k)}$ is the number of times DA $k$ occurred in the data, $\bar{\boldsymbol V}^{(k)}$ is the mean of vectors associated with DA $k$, and $\boldsymbol S_k = \sum_{d_{j,i} = k}{(\boldsymbol v_{j,i} - \bar{\boldsymbol V}^{(k)})(\boldsymbol v_{j,i} - \bar{\boldsymbol V}^{(k)})^\top}$ is scaled form of the covariance of these vectors.
Note that $\kappa$, $\nu$, $\boldsymbol\mu$, and $\boldsymbol\Psi$ are hyper-parameters which need to be set in advance.

Now we can construct the final posterior predictive distribution used for sampling DA assignments:

\noindent
\begin{multline}\label{eq:ctx-gmm}
P(d_{j,i}=k|\boldsymbol D_{\setminus j,i},\boldsymbol V_{\setminus j,i}) \propto \\
\begin{array}{*{50}l}
   &  P(d_{j,i}|d_{j,i-1},\boldsymbol d_{\setminus j,i}) \\
   \times &  P(d_{j,i+1}|d_{j,i}, \boldsymbol d_{\setminus j,i+1} )  \\
   \times &  t_{\nu_k-K+1}(\boldsymbol v_{j,i}|\boldsymbol\mu_k,\frac{\kappa_k+1}{\kappa_k}\boldsymbol\Sigma_k). \\
\end{array}
\end{multline}

\noindent
The product of the first two parts in the equation expresses the score proportional to the probability of DA at position $i$ in the $j$-th dialogue given the surrounding HMM states. The third part expresses the probability of DA assignment given the current feature vector $\boldsymbol v_{j,i}$ and all other DA assignments.

We also present the simplified version of the model that is in fact the standard Gaussian mixture model (GMM). This model does not capture the dependencies between surrounding DAs in the dialogue. Posterior predictive distribution is as follows:

\begin{multline}\label{eq:gmm}
P(d_{j,i}=k|\boldsymbol D_{\setminus j,i},\boldsymbol V_{\setminus j,i}) \propto \frac{n_{\setminus j,i}^{(k)} + \alpha}{N - 1 + K\alpha} \\
\times  t_{\nu_k-K+1}(\boldsymbol v_{j,i}|\boldsymbol\mu_k,\frac{\kappa_k+1}{\kappa_k}\boldsymbol\Sigma_k).
\end{multline}

In Section \ref{sec:xps} we provide comparison of both models to see the strengths of using DA context.

\section{DA Feature Vector\label{sec:features}}

The real-valued vectors $\boldsymbol v_{j,i}$ are expected to represent the meaning of $d_{j,i}$. We use semantic composition approach. It is based on \emph{Frege's principle of compositionality} \cite{Pelletier1994}, which states that the meaning of a complex expression is determined as a composition of its parts, i.e. words. 

We use linear combination of word vectors, where the weights are represented by the inverse-document-frequency (IDF) values of words. We use Global Vectors (GloVe) \cite{pennington2014glove} for word vector representation. We use pre-trained word vectors\footnote{Available at \url{http://nlp.stanford.edu/projects/glove/}.} on 6B tokens from Wikipedia 2014 and Gigaword 5. \newcite{brychcin-svoboda:2016:SemEval} showed that this approach leads to very good representation of short sentences.

For supervised approaches we also use bag-of-words (BoW) representation of an utterance, i.e. separate binary feature representing the occurrence of a word in the utterance.

\section{Experimental Results and Discussion\label{sec:xps}}

\begin{table*}[!ht]
\begin{center}
\resizebox{\textwidth}{!}{
\begin{tabular}{lrrrrrrrr}
\specialrule{.15em}{.1em}{.1em} 
& \bf Model  &  \bf AC & \bf PU & \bf CO & \bf F1 & \bf HO & \bf CM & \bf V1     \\
\hline
\multirow{3}{*}{\rotatebox[origin=c]{90}{Extreme}} 
& Random labels 	&	2.6\% &		31.5\%	 &	4.9\%	&	8.5\%	&	6.8\%	&	4.1\%	&	5.1\%   \\
& Distinct labels 	&	0.0\% &	 	100.0\%	 &	0.9\%	&	1.8\%	&	100.0\%	&	26.9\%	&	42.4\% \\
& Majority label 	&	31.5\%	 &	31.5\% &		100.0\%	&	47.9\%	&	0.0\%	&	100.0\%	&	0.0\%   \\
\hline
\multirow{7}{*}{\rotatebox[origin=c]{90}{Supervised}} 
& 	ME GloVe  ($M=50$)		&	63.2\%	&	63.3\%	&	77.8\%	&	69.8\%	&	41.0\%	&	57.3\%	&	47.8\% \\
&	ME GloVe  ($M=100$)	&	64.1\%	&	64.4\%	&	76.9\%	&	70.1\%	&	43.3\%	&	57.3\%	&	49.3\% \\
&	ME GloVe  ($M=200$)	&	64.8\%	&	65.1\%	&	77.2\%	&	70.6\%	&	43.5\%	&	58.1\%	&	49.7\% \\
&	ME GloVe  ($M=300$)	&	65.6\%	&	65.8\%	&	76.0\%	&	70.6\%	&	45.0\%	&	57.7\%	&	50.5\% \\
& 	ME BoW				&	70.4\%	&	70.7\%	&	76.3\%	&	73.4\%	&	51.0\%	&	62.9\%	&	56.3\% \\
& 	ME BoW + GloVe ($M=300$)		&	71.5\%	&	72.0\%	&	76.0\%	&	74.0\%	&	53.2\%	&	62.9\%&	57.7\% \\
& 	ctx ME BoW + GloVe ($M=300$)	&	\bf 72.9\%	&	73.0\%	&	76.1\%	&	\bf 74.5\%	&	53.9\%	&	64.1\%&	\bf 58.6\% \\
\hline
\multirow{14}{*}{\rotatebox[origin=c]{90}{Unsupervised}} 
&	BHMM \cite{ritter2010unsupervised} 	& /	& 60.3\% & 31.2\% & 41.1\% & 43.1\% & 29.1\% & 34.7\%   \\
&	M4 \cite{paul2012mixed} 			& /	& 44.4\% & 45.9\% & 45.1\% & 19.4\% & 16.9\% & 18.0\%  \\
\cline{2-9}
& 	K-means GloVe ($M=50$)				&/	&57.1\%	& 25.9\%	& 35.6\%	& 39.9\%	& 27.5\%	& 32.6\%  \\
& 	K-means GloVe ($M=100$)			&/	&56.7\%	& 29.5\%	& 38.8\%	& 39.9\%	& 28.9\%	& 33.5\% \\
& 	K-means GloVe ($M=200$)			&/	&56.9\%	& 32.4\%	& 41.3\%	& 39.7\%	& 31.2\%	& 35.0\% \\
& 	K-means GloVe ($M=300$)			&/	&57.4\%	& 31.2\%	& 40.4\%	& 40.2\%	& 30.3\%	& 34.6\% \\
\cline{2-9}
& 	GMM GloVe ($M=50$)		&	/	&54.4\%	& 51.8\%	& 53.1\%	& 34.0\%	& 37.7\%	& 35.8\% \\
& 	GMM GloVe ($M=100$)		&	/	&53.8\%	& 58.1\%	& 55.9\%	& 33.7\%	& 40.0\%	& 36.5\% \\
& 	GMM GloVe ($M=200$)		&	/	&52.1\%	& 76.9\%	& 62.1\%	& 31.3\%	& 43.6\%	& 36.4\% \\
& 	GMM GloVe ($M=300$)		&	/	&52.7\%	& 79.8\%	& 63.5\%	& 30.1\%	& 45.2\%	& 36.1\% \\
\cline{2-9}
& 	ctx GMM GloVe ($M=50$)		&	/	&55.1\%	& 60.0\%	& 57.5\%	& 36.4\%	& 42.4\%	& 39.1\% \\
& 	ctx GMM GloVe ($M=100$)	&	/	&53.8\%	& 81.7\%	& 64.9\%	& 32.3\%	& 51.7\%	& 39.8\% \\
& 	ctx GMM GloVe ($M=200$)	&	/	&54.7\%	& 81.4\%	& 65.5\%	& 32.1\%	& 51.9\%	& 39.7\% \\
& 	ctx GMM GloVe ($M=300$)	&	/	&55.2\%	& 81.0\%	& \bf 65.7\% & 34.4\%	& 51.4\%	& \bf 41.2\% \\
\specialrule{.15em}{.1em}{.1em} 
\end{tabular}
}
\end{center}
\caption{\label{tab:results} Accuracy (AC), purity (PU), collocation (CO), f-measure (F1), homogeneity (HO), completeness (CM), and v-measure (V1) for proposed models expressed in percents.}
\end{table*}

We use Switchboard-DAMSL corpus \cite{Jurafsky97a} to evaluate the proposed methods. The corpus contains transcriptions of telephone conversations between multiple speakers that do not know each other and are given a topic for discussion. 
We adopt the same set of 42 DA labels and the same train/test data split as suggested in \cite{Stolcke00}\footnote{1115 dialogues (196,258 utterances) are used for training while 19 dialogues (4186 utterances) for testing. More information about the data split can be found at \url{http://web.stanford.edu/~jurafsky/ws97}.}.

In our experiments we set $\kappa = 0$ , $\boldsymbol\mu = \boldsymbol 0$, $\nu = K$, $\boldsymbol \Psi = \mathds{1}$, and $\alpha = 50/K$. These parameters are recommended by \cite{NAS:2004:griffiths,Murphy:2012} and we also confirm them empirically. We always perform 1000 iterations of Gibbs sampling. The number of clusters (mixture size) is $K = 42$. The dimension of GloVe vectors ranges between $M=50$ and $M=300$.

DA induction task is in fact the clustering problem. We cluster DA utterances and we assign the same label to utterances within one cluster. Standard metrics for evaluating quality of clusters are \emph{purity} (PU), \emph{collocation} (CO), and their harmonic mean (F1). In the last years, \emph{v-measure} (V1) have also become popular. This entropy-based measure is defined as harmonic mean between \emph{homogeneity} (HO -- the precision analogue) and \emph{completeness} (CM -- the recall analogue). \newcite{rosenberg-hirschberg:2007} presents definition and comparison of all these metrics. Note the same evaluation procedure  is often used for different clustering tasks, e.g., unsupervised part-of-speech induction \cite{christodoulopoulos-goldwater-steedman:2010} or unsupervised semantic role labeling \cite{woodsend-lapata:2015}.

Table \ref{tab:results} presents the results of our experiments. We compare both supervised and unsupervised approaches. Models incorporating the information about surrounding DAs (context) are denoted by prefix \emph{ctx}. We show the results of three unsupervised approaches: K-means clustering, GMM without context (Eq. \ref{eq:gmm}), and context-dependent GMM (Eq. \ref{eq:ctx-gmm}). We use Maximum Entropy (ME) classifier \cite{Berger:1996} for the supervised approach. For the context-dependent version we perform two-round classification: firstly, without the context information and secondly, incorporating the output from the previous round.

In addition, Table \ref{tab:results} provides results for the three extreme cases: \emph{random label}, \emph{majority label}, and \emph{distinct label} for each utterance (a single utterance per cluster). Note the last mentioned achieved v-measure of 42.4\%. In this case, however, completeness approaches 0\% with the rising size of the test data (so v-measure does too). So this number cannot be taken into account.

To the best of our knowledge, the best performing supervised system on Switchboard-DAMSL corpus is presented in \cite{kalchbrenner-blunsom:2013} and achieves accuracy of 73.9\%. Our best supervised baseline is approximately 1\% worse. In all experiments the context information proved to be very useful. The best result among unsupervised models is achieved with 300-dimensional GloVe (F1 score 65.7\% and v-measure 41.2\%). We  outperform both the block HMM (BHMM) \cite{ritter2010unsupervised} achieving F1 score 41.1\% and v-measure 34.7\% and mixed-membership HMM (M4) \cite{paul2012mixed} achieving F1 score 45.1\% and v-measure 18.0\%\footnote{Both implementations are available at \url{http://cmci.colorado.edu/~mpaul/downloads/mm.php}. We use recommended settings. Note the comparison with M4 is not completely fair, because it does not directly assign DAs to utterances (instead, it assigns DAs to each token). We always took the most frequent token DA in utterance as final DA.}. If we compare our method with the supervised version (F1 score 74.5\% and v-measure 58.6\%) we can state that HMM with GMMs is very promising direction for the unsupervised DA induction task.

\section{Conclusion and Future Work\label{sec:conclusion}}

We introduced HMM based model for unsupervised DA induction. We represent each utterance as a real-valued vector encoding the meaning. Our model predicts these vectors in the context of DA utterances. We compared our model with several strong baselines and showed its strengths. Our Java implementation is available for research purposes at \url{https://github.com/brychcin/unsup-dial-act-induction}.

As the main direction for future work, we plan to experiment with more languages and more corpora.
Also, more thorough study of feature vector representation should be done.

We plan to investigate the learning process much more deeply. It was beyond the scope of this paper to evaluate the time expenses of the algorithm. Moreover, there are several possibilities how to speed up the process of parameter estimation, e.g. by Cholesky decomposition of the covariance matrix as described in \cite{das-zaheer-dyer:2015}. In our current implementation the number of DAs is set in advance. It could be very interesting to use non-parametric version of GMM, i.e. to change the sampling scheme to estimate the number of DAs by Chinese restaurant process.

\section*{Acknowledgments}
This publication was supported by the project LO1506 of the Czech Ministry of Education, Youth and Sports. Computational resources were provided by the CESNET LM2015042 and the CERIT Scientific Cloud LM2015085, provided under the programme "Projects of Large Research, Development, and Innovations Infrastructures". Lastly, we would like to thank the anonymous reviewers for their insightful feedback.

\FloatBarrier

\bibliography{paper}

\begin{thebibliography}{}

\bibitem[\protect\citename{Allen \bgroup et al.\egroup }2007]{Allen:2007}
James Allen, Nathanael Chambers, George Ferguson, Lucian Galescu, Hyuckchul
  Jung, Mary Swift, and William Taysom.
\newblock 2007.
\newblock Plow: A collaborative task learning agent.
\newblock In {\em Proceedings of the 22Nd National Conference on Artificial
  Intelligence - Volume 2}, AAAI'07, pages 1514--1519. AAAI Press.

\bibitem[\protect\citename{Berger \bgroup et al.\egroup }1996]{Berger:1996}
Adam~L. Berger, Vincent J.~D. Pietra, and Stephen A.~D. Pietra.
\newblock 1996.
\newblock A maximum entropy approach to natural language processing.
\newblock {\em Computational Linguistics}, 22:39--71, March.

\bibitem[\protect\citename{Brychc\'{i}n and
  Svoboda}2016]{brychcin-svoboda:2016:SemEval}
Tom\'{a}\v{s} Brychc\'{i}n and Luk\'{a}\v{s} Svoboda.
\newblock 2016.
\newblock {UWB} at {S}em{E}val-2016 {T}ask 1: {S}emantic {T}extual {S}imilarity
  using {L}exical, {S}yntactic, and {S}emantic {I}nformation.
\newblock In {\em Proceedings of the 10th International Workshop on Semantic
  Evaluation (SemEval-2016)}, pages 588--594, San Diego, California, June.
  Association for Computational Linguistics.

\bibitem[\protect\citename{Christodoulopoulos \bgroup et al.\egroup
  }2010]{christodoulopoulos-goldwater-steedman:2010}
Christos Christodoulopoulos, Sharon Goldwater, and Mark Steedman.
\newblock 2010.
\newblock Two decades of unsupervised {POS} induction: How far have we come?
\newblock In {\em Proceedings of the 2010 Conference on Empirical Methods in
  Natural Language Processing}, pages 575--584, Cambridge, MA, October.
  Association for Computational Linguistics.

\bibitem[\protect\citename{Crook \bgroup et al.\egroup
  }2009]{crook2009unsupervised}
Nigel Crook, Ramon Granell, and Stephen Pulman.
\newblock 2009.
\newblock Unsupervised classification of dialogue acts using a {D}irichlet
  process mixture model.
\newblock In {\em Proceedings of the SIGDIAL 2009 Conference}, pages 341--348,
  London, UK, September. Association for Computational Linguistics.

\bibitem[\protect\citename{Das \bgroup et al.\egroup
  }2015]{das-zaheer-dyer:2015}
Rajarshi Das, Manzil Zaheer, and Chris Dyer.
\newblock 2015.
\newblock Gaussian lda for topic models with word embeddings.
\newblock In {\em Proceedings of the 53rd Annual Meeting of the Association for
  Computational Linguistics and the 7th International Joint Conference on
  Natural Language Processing (Volume 1: Long Papers)}, pages 795--804,
  Beijing, China, July. Association for Computational Linguistics.

\bibitem[\protect\citename{Diehl \bgroup et al.\egroup }2007]{Diehl:2007}
Christopher~P. Diehl, Galileo Namata, and Lise Getoor.
\newblock 2007.
\newblock Relationship identification for social network discovery.
\newblock In {\em Proceedings of the 22nd National Conference on Artificial
  Intelligence}, AAAI'07, pages 546--552. AAAI Press.

\bibitem[\protect\citename{Griffiths and Steyvers}2004]{NAS:2004:griffiths}
Thomas~L. Griffiths and Mark Steyvers.
\newblock 2004.
\newblock {Finding scientific topics}.
\newblock {\em Proceedings of the National Academy of Sciences of the United
  States of America}, 101(Suppl 1):5228--5235, April.

\bibitem[\protect\citename{Joty \bgroup et al.\egroup
  }2011]{joty2011unsupervised}
Shafiq Joty, Giuseppe Carenini, and Chin-Yew Lin.
\newblock 2011.
\newblock Unsupervised modeling of dialog acts in asynchronous conversations.
\newblock In {\em Proceedings of the 22nd International Joint Conference on
  Artificial Intelligence}, IJCAI'11, pages 1807--1813. AAAI Press.

\bibitem[\protect\citename{Jurafsky and Martin}2009]{Jurafsky:2009}
Daniel Jurafsky and James~H. Martin.
\newblock 2009.
\newblock {\em Speech and Language Processing (2Nd Edition)}.
\newblock Prentice-Hall, Inc., Upper Saddle River, NJ, USA.

\bibitem[\protect\citename{Jurafsky \bgroup et al.\egroup }1997]{Jurafsky97a}
Daniel Jurafsky, Elizabeth Shriberg, and Debra Biasca.
\newblock 1997.
\newblock Switchboard {SWBD-DAMSL} {S}hallow-{D}iscourse-{F}unction
  {A}nnotation ({C}oders {M}anual, {D}raft 13).
\newblock Technical Report 97-01, University of Colorado, Institute of
  Cognitive Science.

\bibitem[\protect\citename{Kalchbrenner and
  Blunsom}2013]{kalchbrenner-blunsom:2013}
Nal Kalchbrenner and Phil Blunsom.
\newblock 2013.
\newblock Recurrent convolutional neural networks for discourse
  compositionality.
\newblock In {\em Proceedings of the Workshop on Continuous Vector Space Models
  and their Compositionality}, pages 119--126, Sofia, Bulgaria, August.
  Association for Computational Linguistics.

\bibitem[\protect\citename{Kl\"{u}wer \bgroup et al.\egroup
  }2010]{kluwer2010using}
Tina Kl\"{u}wer, Hans Uszkoreit, and Feiyu Xu.
\newblock 2010.
\newblock Using syntactic and semantic based relations for dialogue act
  recognition.
\newblock In {\em Coling 2010: Posters}, pages 570--578, Beijing, China,
  August. Coling 2010 Organizing Committee.

\bibitem[\protect\citename{Murphy}2012]{Murphy:2012}
Kevin~P. Murphy.
\newblock 2012.
\newblock {\em Machine Learning: A Probabilistic Perspective}.
\newblock The MIT Press.

\bibitem[\protect\citename{Paul}2012]{paul2012mixed}
Michael~J. Paul.
\newblock 2012.
\newblock Mixed membership markov models for unsupervised conversation
  modeling.
\newblock In {\em Proceedings of the 2012 Joint Conference on Empirical Methods
  in Natural Language Processing and Computational Natural Language Learning},
  pages 94--104, Jeju Island, Korea, July. Association for Computational
  Linguistics.

\bibitem[\protect\citename{Pelletier}1994]{Pelletier1994}
Francis~Jeffry Pelletier.
\newblock 1994.
\newblock The principle of semantic compositionality.
\newblock {\em Topoi}, 13(1):11--24.

\bibitem[\protect\citename{Pennington \bgroup et al.\egroup
  }2014]{pennington2014glove}
Jeffrey Pennington, Richard Socher, and Christopher Manning.
\newblock 2014.
\newblock Glove: Global vectors for word representation.
\newblock In {\em Proceedings of the 2014 Conference on Empirical Methods in
  Natural Language Processing (EMNLP)}, pages 1532--1543, Doha, Qatar, October.
  Association for Computational Linguistics.

\bibitem[\protect\citename{Ritter \bgroup et al.\egroup
  }2010]{ritter2010unsupervised}
Alan Ritter, Colin Cherry, and Bill Dolan.
\newblock 2010.
\newblock Unsupervised modeling of twitter conversations.
\newblock In {\em Human Language Technologies: The 2010 Annual Conference of
  the North American Chapter of the Association for Computational Linguistics},
  pages 172--180, Los Angeles, California, June. Association for Computational
  Linguistics.

\bibitem[\protect\citename{Rosenberg and
  Hirschberg}2007]{rosenberg-hirschberg:2007}
Andrew Rosenberg and Julia Hirschberg.
\newblock 2007.
\newblock {V}-measure: A conditional entropy-based external cluster evaluation
  measure.
\newblock In {\em Proceedings of the 2007 Joint Conference on Empirical Methods
  in Natural Language Processing and Computational Natural Language Learning
  (EMNLP-CoNLL)}, pages 410--420, Prague, Czech Republic, June. Association for
  Computational Linguistics.

\bibitem[\protect\citename{Seo \bgroup et al.\egroup }2009]{Seo:2009}
Jangwon Seo, W.~Bruce Croft, and David~A. Smith.
\newblock 2009.
\newblock Online community search using thread structure.
\newblock In {\em Proceedings of the 18th ACM Conference on Information and
  Knowledge Management}, CIKM '09, pages 1907--1910, New York, NY, USA. ACM.

\bibitem[\protect\citename{Stolcke \bgroup et al.\egroup }2000]{Stolcke00}
Andreas Stolcke, Klaus Ries, Noah Coccaro, Elizabeth Shriberg, Rebecca Bates,
  Daniel Jurafsky, Paul Taylor, Rachel Martin, Carol~Van Ess-Dykema, and Marie
  Meteer.
\newblock 2000.
\newblock Dialog act modeling for automatic tagging and recognition of
  conversational speech.
\newblock In {\em Computational Linguistics}, volume~26, pages 339--373.

\bibitem[\protect\citename{Wang \bgroup et al.\egroup }2011]{wang-EtAl:2011}
Li~Wang, Marco Lui, Su~Nam Kim, Joakim Nivre, and Timothy Baldwin.
\newblock 2011.
\newblock Predicting thread discourse structure over technical web forums.
\newblock In {\em Proceedings of the 2011 Conference on Empirical Methods in
  Natural Language Processing}, pages 13--25, Edinburgh, Scotland, UK., July.
  Association for Computational Linguistics.

\bibitem[\protect\citename{Woodsend and Lapata}2015]{woodsend-lapata:2015}
Kristian Woodsend and Mirella Lapata.
\newblock 2015.
\newblock Distributed representations for unsupervised semantic role labeling.
\newblock In {\em Proceedings of the 2015 Conference on Empirical Methods in
  Natural Language Processing}, pages 2482--2491, Lisbon, Portugal, September.
  Association for Computational Linguistics.

\end{thebibliography}
\bibliographystyle{eacl2017}

\end{document}